\documentclass[10pt]{article}
\usepackage[preprint]{tmlr}

\usepackage{amsmath,amsfonts,bm}

\def\eqref#1{equation~\ref{#1}}

\def\1{\bm{1}}

\DeclareMathAlphabet{\mathsfit}{\encodingdefault}{\sfdefault}{m}{sl}
\SetMathAlphabet{\mathsfit}{bold}{\encodingdefault}{\sfdefault}{bx}{n}

\usepackage{hyperref}
\usepackage{graphicx}
\usepackage{subcaption}
\usepackage{url}
\usepackage{placeins}
\usepackage{xcolor}
\usepackage{bbm}
\usepackage{mathtools}
\usepackage{dsfont}

\definecolor{myblue}{rgb}{0.141, 0.514, 0.596}
\definecolor{mypurple}{rgb}{0.494, 0.341, 0.761}
\definecolor{myorange}{rgb}{0.851, 0.510, 0.098}

\title{Stagewise and Interpretable Emergence of Inside--Outside Computation in Transformer-based Language Models}

\author{
Jack T. Parley \\
Institute of Physics \\
\'{E}cole Polytechnique F\'{e}d\'{e}rale de Lausanne (EPFL) \\
Lausanne, Switzerland \\
\texttt{jack.parley@epfl.ch} \quad \texttt{jparley4d@gmail.com}
}

\newcommand{\tell}{\tilde{L}}

\def\month{MM}
\def\year{YYYY}
\def\openreview{\url{https://openreview.net/forum?id=XXXX}}

\begin{document}

\maketitle
\begin{abstract}
    Variability in syntactic structure and parsing ambiguity are fundamental aspects of natural language which modern LLMs have to learn to deal with proficiently. Previous ``post-mortem" studies have shown that fully trained Transformers indeed approximate canonical parsing algorithms such as inside-outside, but little is established theoretically about how this is actually learned. Here, we make a first step towards addressing this by studying the learning dynamics in synthetic context-free grammars with variable parse tree topology. The stagewise learning curves are fully interpretable, in the sense that all of the loss stages a) can be computed analytically, and b) can be clearly associated to the representation of a hidden latent tree variable, which allows to lift ambiguity and utilize information from wider contexts. By probing representations, we visualize the emergence of inside-outside computation across depth and during training: while during early stages local (bigram/trigram) information is used and involves only a few initial layers, gradually the Transformer develops an encoder-decoder structure where the most abstract latent information is encoded in the middle, and the latter half of the network is used to progressively refine the final token prediction. This organization of computation is reminiscent of findings in real-world LLMs, and importantly applies here to a synthetic model which allows full analytical control of the data statistics, thus opening the door to theoretical progress. 
\end{abstract}

\begin{figure*}[!t]
    \centering





        \begin{subfigure}[t]{0.42\textwidth}
        \centering
        \includegraphics[
            width=\linewidth,
            height=0.25\textheight,
            keepaspectratio
        ]{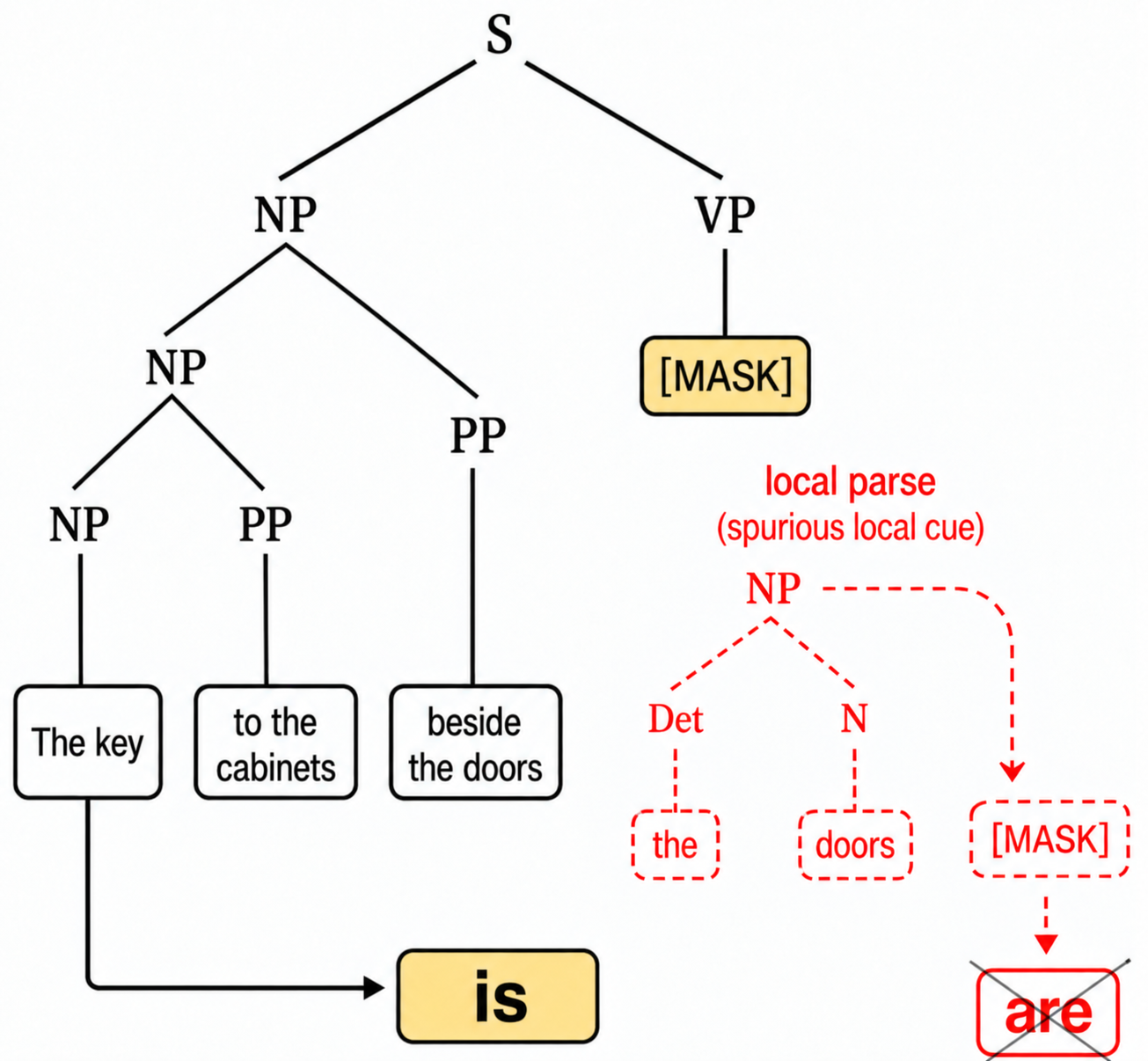}
        \caption{Inspiration: natural language syntax}
        \label{fig:syntax_example}
    \end{subfigure}
    \hfill
    \begin{subfigure}[t]{0.54\textwidth}
        \centering
        \includegraphics[
            width=\linewidth,
            height=0.25\textheight,
            keepaspectratio
        ]{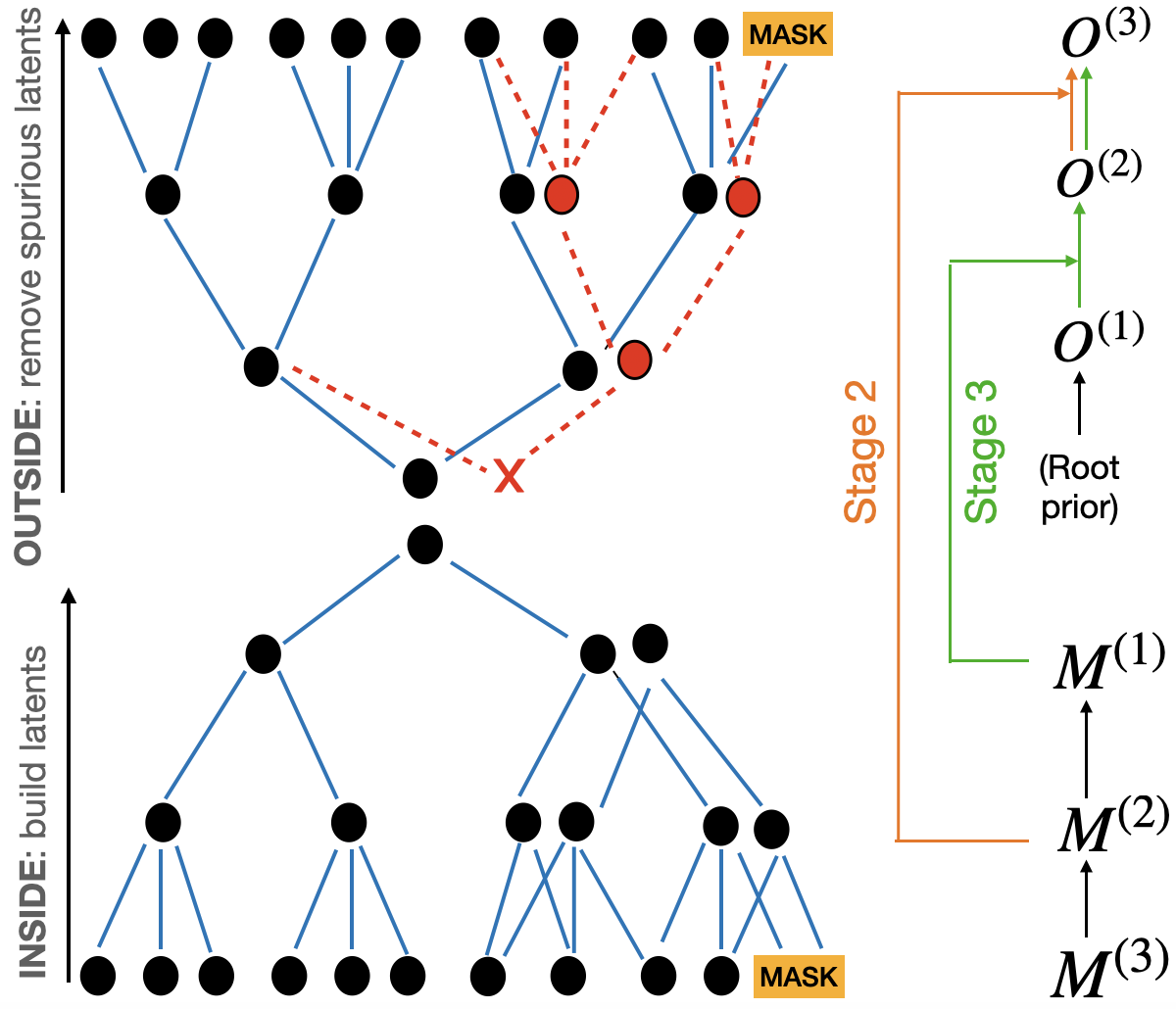}
        \caption{Inside-outside computation in the VT-RHM.}
        \label{fig:IO_sketch}
    \end{subfigure}

    \vspace{0.8em}
    \begin{subfigure}[t]{0.8\textwidth}
        \centering
        \includegraphics[
            width=\linewidth,
        ]{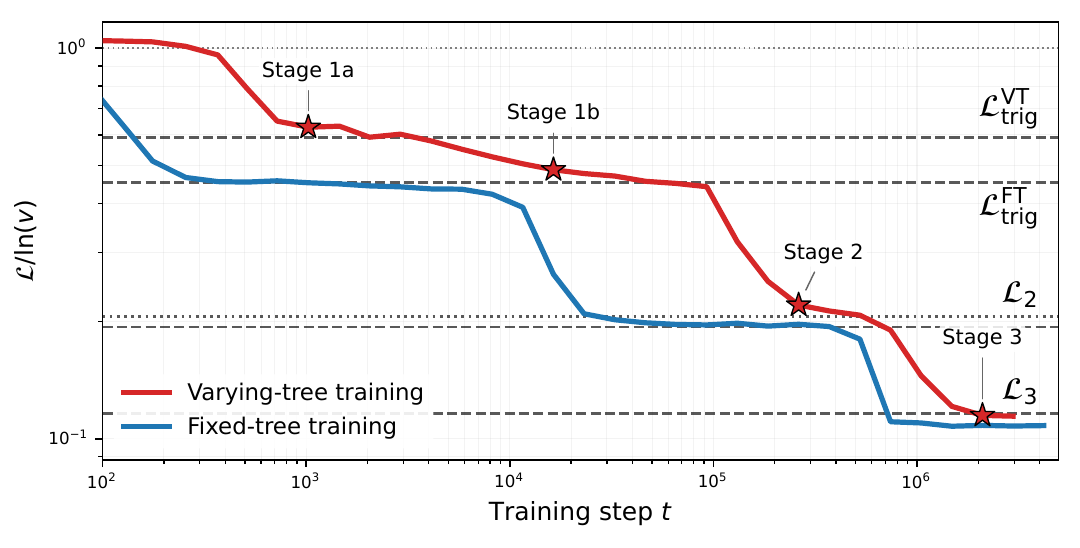}
        \caption{Stagewise learning curves of GPT-2 style Transformers.}
        \label{fig:training_curves}
    \end{subfigure}

    \caption{
    \textbf{Inside--outside computation and stagewise learning.}
    (a) Example illustrating the importance of hierarchical structure in natural language, where attending to local cues would lead to an erroneous completion. (b) Illustration of the inside-outside algorithm at work in a depth $L=3$ VT-RHM. While building the hidden latent structure (inside pass), due to local ambiguity spurious nonterminals are found, which would lead to erroneous predictions. This has to be ``cleaned up” by an outside pass: in the case shown, the alternative tree topology (in red) is found to be spurious due to its grammatical incompatibility with the first half of the sentence (the top rule cannot be completed). Mathematically, the algorithm proceeds by building up inside tensors $M^{(l)}$, then progressively integrating this information into outside tensors $O^{(l)}$ (see text). The case sketched corresponds to Stage 3, as it uses all layers of rules. (c) Test cross-entropy loss on the ternary-tree sentence completion task, for GPT-2 style Transformers trained on fixed-tree or varying-tree training sets (see Fig.~\ref{fig:varyig_vs_fixed}). Horizontal plateaus indicate the theoretical loss values studied in Sec.~\ref{sec:stages}. The stars refer to the checkpoints used for the probing measurements, chosen to be representative of each learning stage (their precise location within each stage does not affect results qualitatively).
    }
    \label{fig:training_and_IO_combined}
\end{figure*}

\section{Introduction}

Natural language is syntactically \textit{ambiguous}. Grammars have evolved to optimize efficient communication while balancing complexity and ambiguity~\citep{hahn_universals_2020,gibson2019efficiency}, but as a result many commonplace linguistic constructions are inherently problematic. A classic example in English is prepositional phrase attachment: the phrase ``I saw him with the binoculars", has two equally valid interpretations, each of which corresponds to a distinct parse tree. This is an example of unresolvable ambiguity, at least at the sentence level. For an LLM predicting the next token, there are plenty of examples where the sentence may be locally problematic, but where this can be resolved given sufficient grammar knowledge. For example, in the simple example of Subject-Verb agreement sketched in Fig.~\ref{fig:syntax_example}, attending to any of the closest nouns leads to a wrong prediction; if the entire sentence is parsed correctly, however, the subject is identified and the spurious local cues cease to be problematic.

A substantial body of work has investigated how modern language models acquire and represent syntactic structure. Many works have emphasized how Transformers struggle with syntactic generalization~\citep{yao2022structural}, while others have shown the rich representation of syntax by probing activations in state-of-the art models~\citep{manning2020emergent}. Studying such questions cleanly ``in the wild", however, is notoriously difficult. Although some successes have been reported in identifying specific syntax-related circuits~\citep{ferrando2024similarity}, recovering and interpreting such circuits more generally remains challenging, often requiring manual intervention and human judgment~\citep{ameisen2025circuit,lindsey2025biology}. 

Mathematically, if we model language with a probabilistic-context-free-grammar (PCFG), the question can be phrased as whether and how LLMs learn canonical parsing algorithms to deal with syntax. PCFGs, originating in formal language theory~\citep{chomsky_syntactic_1957}, simply define a set of grammar rules with a corresponding probability. In computational linguistics, the associated canonical algorithm for dealing with syntactic ambiguity is the inside-outside algorithm (I-O)~\citep{baker_trainable_1979}. From an inference perspective, this constitutes the Bayes-optimal algorithm to perform masked token prediction given data generated from a PCFG~\citep{zhao_Transformers_2023}.

Some previous works have attempted to ask precisely the question of whether and how inside-outside is implemented in Transformer~\citep{vaswani_attention_2017}-based language models, mainly~\citep{allen-zhu_physics_2023,zhao_Transformers_2023}. \citep{allen-zhu_physics_2023}, in particular, performed ``post-mortem" interpretability analysis on the attention patterns of the language model, showing that it mimics dynamic programming (i.e. inside-outside), but sheds little light on the learning dynamics or the organization of latent information and outside computation across depth. In a recent paper, \citep{zhou2026hierarchies} does study latent structure across depth, but the experiments were limited to (next-) token prediction on a fixed-tree topology~\footnote{We refer here also to the work of~\citep{garnier-brun_how_2025}, which also analyzed Transformer computation but for fixed-tree topology.}. Finally, and perhaps most importantly, although all these works consider some form of synthetic grammar none provide any analytical/quantitative control over the generative process, i.e. the data statistics induced by the grammar or associated language.

As an initial step towards filling this gap, we study here the learning dynamics of a family of synthetic PCFGs introduced by~\citep{parley2026deep}. The key ingredients of this model, which is an extension of the model of~\citep{cagnetta_how_2024} to varying-tree topologies (thus allowing for syntactic ambiguity), are twofold: it is layered, and the grammar is built by sampling rules uniformly without replacement from a finite pool of nonterminal symbols. These two ingredients are key to opening the door for interpretability: the randomness of the rules allows for analytical calculations, while, together with the layered structure, a hierarchy of correlation strengths is built into the grammar, in turn inducing a stagewise interpretable learning of the token-prediction task.

Building on this interpretable stagewise learning, we will show that, with these very simple ingredients, our model of data is rich enough to recover some salient features of the way language computation has been found to be organized in LLMs. While during the early stages of training purely local statistics (bigram/trigram) is used and only the initial layers of the Transformer are active, gradually the Transformer develops an encoder-decoder structure where the deepest latent information is encoded in the middle, and the latter half of the network is used to progressively refine the final prediction. This echoes the general picture which has emerged empirically regarding the functional differentiation of layers in Transformer language models: early layers tend to encode local and surface-level information, while intermediate and deeper layers progressively integrate~\footnote{\citep{kervadec2026tracing} specifically showed, by tracing computation density in state of the art LLMs, that the effect of later layers is dominated by attention heads and corresponds to a progressive refinement of the final output.} more abstract, contextual, and hierarchical representations~\citep{tenney2019bert,jawahar2019bert,kervadec2026tracing}.

Although inside-outside computation clearly does not account for all of natural language processing, what we wish to emphasize is that we recover this rich behaviour in a synthetic model which allows for complete analytical/quantitative control over all statistics, and in particular allows to compute all correlations relevant for learning. This was done in~\citep{parley2026deep} already in the context of supervised learning. We will thus argue that this data model can potentially open the door to building quantitative theories of many interesting phenomena related to syntax and parsing in LLMs.


\section{Varying-tree RHM}\label{sec:VT-RHM}

We give here only a brief and informal presentation of our family of synthetic grammars, which was presented formally and in great detail in~\citep{parley2026deep}. The Varying-tree Random Hierarchy Model (VT-RHM), defines an ensemble of layered PCFGs of depth $L$. To build a single grammar instance $\mathcal{G}$, we introduce as
\begin{equation}
\mathcal{N}^{(0)},\mathcal{N}^{(1)},\ldots,\mathcal{N}^{(L)}
\end{equation}
the sets of symbols at successive levels of the hierarchy, with
$\mathcal{N}^{(L)}\equiv\Sigma$ the terminal vocabulary. Throughout, we take
\begin{equation}
\left|\mathcal{N}^{(\ell)}\right|=v,
\qquad
\ell=0,\ldots,L,
\end{equation}
so that every level contains $v$ possible symbols.

For each level $\ell=1,\ldots,L$ and each parent symbol
$z\in\mathcal{N}^{(\ell-1)}$, the grammar contains $m_2$ binary rules
\begin{equation}
z\to(a,b),
\qquad
a,b\in\mathcal{N}^{(\ell)},
\end{equation}
and $m_3$ ternary rules
\begin{equation}
z\to(a,b,c),
\qquad
a,b,c\in\mathcal{N}^{(\ell)}.
\end{equation}
For every parent $z$, the grammatical pairs $(a,b)$ and triples $(a,b,c)$
are sampled uniformly without replacement from the $v^2$ and $v^3$
possible combinations, respectively. 

We stress that, conceptually, we are explaining how a single grammar instance is sampled; once the grammar is given, one then can sample sentences (strings of tokens) from the grammar by expanding each nonterminal symbol according to one of its associated rules. This process starts by selecting (with uniform probability $1/v$) a root symbol from $\mathcal{N}^{(0)}$, and expanding across $L$ layers until a string of terminal symbols is produced. For simplicity, we consider these rule probabilities to be uniform, i.e. each binary rule has a probability $p_2/m_2$, and each ternary ternary rule $p_3/m_3$, with $p_2+p_3=1$ so that the PCFG is normalized. $\{p_2,p_3\}$, therefore, set the overall probability of binary vs ternary branchings. In general, the sentence length $d$ will fluctuate in the integer interval $[2^L,\dots 3^L]$.

The final important parameter to define is $f$, which we use to parametrize the number of rules following $m_2=fv$ and $m_3=fv^2$. Importantly, $f$ will also control the level of local ambiguity of a grammar instance. To see this, notice that $f$ corresponds to the average fraction of grammatical binary/ternary tuples (i.e., that can be derived by some parent symbol) with respect to the total set of $v^2$, $v^3$ distinct tuples. Therefore, if $(a,b,c)$ is chosen as a ternary rule, both $(a,b)$ and $(b,c)$ can also be present in the grammar with probability $f$. If, in the final sequence, they are by chance followed or preceded by the appropriate symbol ($c$ or $a$), $(a,b,c)$ will appear fortuitously, causing ambiguity vis-\`{a}-vis the genuine occurrence of $(a,b,c)$ as a ternary rule.

\subsection{Curse of dimensionality and combinatorial explosion of tree topologies}

Having defined a grammar instance $\mathcal{G}$, the associated language is the set of all terminal strings $\mathbf{x}$ such that $P_{\mathcal{G}}(\mathbf{x})>0$,  where this probability is obtained by summing over all possible derivations of $\mathbf{x}$, each weighted by the product of the probabilities of the production rules on the tree. The curse of dimensionality~\citep{bishop2006pattern} is illustrated here by the fact that the size of this language grows exponentially in the sentence length, hence doubly exponentially with depth $L$.

To get a sense of numbers, for the grammar parameters considered in the experiments below ($L=3$, $v=8$, $m_3=16$), just considering the tree topology with only ternary branchings yields a total number of sentences $vm_3^{13}\sim 10^{16}$, orders of magnitude beyond any training set size we can realistically provide to the Transformer. Learning a self-supervised task such as predicting the last token (see below), therefore, will only succeed in a sample-efficient manner if compositional generalization occurs, i.e. if the Transformer exploits the hidden latent structure of the grammar. Besides the number of generated sentences, the number of distinct tree topologies also grows doubly exponentially with depth: for $L=3$, this number~\footnote{At depth \(L=2\), the numbers of distinct topologies for sequence lengths \(d=4,\ldots,9\) are \(N_t(d,2)=(1,2,2,3,3,1)\) (see Fig.10 in~\citep{parley2026deep}). Since each leaf can independently branch into either two or three children, the total number at depth \(L=3\) is \(\sum_{d=4}^{9}2^dN_t(d,2)=1872\).} is already $N_t=1872$. Efficiently accounting for such a large number of possible tree topologies requires a strategy that avoids explicit enumeration, as provided by the inside-outside dynamic programming algorithm.




\section{Learning task: sentence completion (last-token prediction)}

To study the acquisition of inside-outside, we will consider a self-supervised sentence completion (last-token prediction) task. That is, given a sentence prefix $x_1,\dots x_{d-1}$, predict $x_d$. No information is given regarding the hidden latent variables. Last-token prediction can be seen as a special sub-case of a more general next-token prediction task; we focus on it here due to its increased interpretability properties.

\subsection{Inside-outside algorithm}

Given a PCFG, the inside-outside (I-O) algorithm~\citep{baker_trainable_1979} is Bayes-optimal for masked-token prediction~\citep{zhao_Transformers_2023}. We provide the full details of the tensorial implementation of I-O for the VT-RHM in App.~\ref{app:I-O}, sketching here only the main idea of the algorithm. Schematically, for a given sentence the inside pass recursively builds up probabilities of substrings conditioned on an ``anchored" nonterminal variable. The substring is defined by its span, with start position $i=1,\dots d$ and span length $\lambda$. The corresponding inside tensor thus contains the inside probabilities
\begin{equation}
M^{(\ell)}_{i,\lambda}[z]
=
P\!\left(
x_i,\ldots,x_{i+\lambda-1}
\,\middle|\,
z
\right),
\end{equation}
that is, the probability that the observed substring of length $\lambda$ starting at position $i$ is generated by the latent symbol $z\in \mathcal{N}^{(l)}$. Note that the inside tensors $\{M^{(l)}\}$ inherit the layered structure of the grammar, and are built bottom up starting from $l=L$ (token level), see also the sketch~\footnote{As one may notice from the sketch, the root-level inside tensor $M^{(0)}$ does not need to be built. This is because, for the purpose of token prediction, our top layer made up of $v$ root symbols with $m_2+m_3$ rules each is in fact equivalent to a single root symbol with $v(m_2+m_3)$ rules attached to it. More precisely, the first outside tensor $O^{(1)}$ can be built directly from the entries of $M^{(1)}$ by marginalizing over the root symbols, see App.~\ref{app:I-O}.} in Fig.~\ref{fig:IO_sketch}. In our context of last-token prediction, we initialize the token-level tensor as $M^{(L)}_{i,1}[z]=\delta_{x_i,z}$ for $i=1,\dots d-1$, and uniform $M^{(L)}_{d,1}[z]=1/v, \ \forall z$ for the final masked position.

To obtain the Bayes-optimal prediction of the masked token, one then performs an outside pass, computing (from root to leaves) quantities of the form
\begin{equation}
O^{(\ell)}_{i,\lambda}[z]
=
P\!\left(
\bm{x}_{\setminus(i,\lambda)},\, z
\right),
\end{equation}
The notation $\bm{x}_{\setminus(i,\lambda)}$ indicates the surrounding ``context" of the span $(i,\lambda)$, i.e. the rest of the sentence excluding $x_i,\ldots,x_{i+\lambda-1}$. The outside tensor thus aggregates information from surrounding context allowing to refine (and correct) what was found in the inside pass. Ultimately, for last-token prediction one is interested in the quantity $O^{(L)}_{d,1}[z]$, from which one can derive the distribution~\footnote{Here and in the following, we denote the random variables corresponding to nonterminal/terminal symbols with capital letters, using lower case for specific realizations.}
\begin{equation}\label{eq:final_IO}
P(X_d=z\mid x_1,\ldots,x_{d-1})
=
\frac{
O^{(L)}_{d,1}[z]
}{
\sum_{z'\in\mathcal{N}^{(L)}}
O^{(L)}_{d,1}[z']
},
\end{equation}
With this, one obtains the Bayes-optimal last-token predictor. 
To characterize our datasets, we average over sentences and consider the conditional
last-token entropy of the distribution induced by a given
grammar instance $\mathcal{G}$,
\begin{equation}\label{eq:Bayes_optimal_loss}
\mathcal{L}_{\mathrm{opt}}
=
-\mathbb{E}_{\mathbf{X}\sim P_{\mathcal{G}}}
\left[
\ln P_{\mathcal{G}}(X_d\mid X_1,\ldots,X_{d-1})
\right]
=
H_{\mathcal{G}}(X_d\mid X_1,\ldots,X_{d-1}),
\end{equation}
which constitutes the information-theoretic lower bound
on the cross-entropy loss of any last-token
predictor.

We finally note that in (\ref{eq:Bayes_optimal_loss}) we have dropped the dependence of $\mathcal{L}_{\mathrm{opt}}$ on the grammar instance. This is because our data model displays self-averaging~\citep{parley2026deep}, so that quantities such as $\mathcal{L}_{\mathrm{opt}}$ measured on different grammar instances concentrate around their mean over grammar realizations $\mathop{\mathbb{E}}_{\mathcal{G}}\left[\mathcal{L}_{\mathrm{opt}}\right]$, even for moderate values of vocabulary size such as $v=8$ considered below. This property, which is the same that allows to compute typical thermodynamic properties of disordered systems in physics~\citep{nishimori2001statistical}, will be crucial also here as it will allow us to understand any single grammar instance in terms of a typical (ensemble-averaged) grammar instance.

\subsection{Our setup}

\begin{figure}[!]
    \centering
    \includegraphics[width=0.6\linewidth]{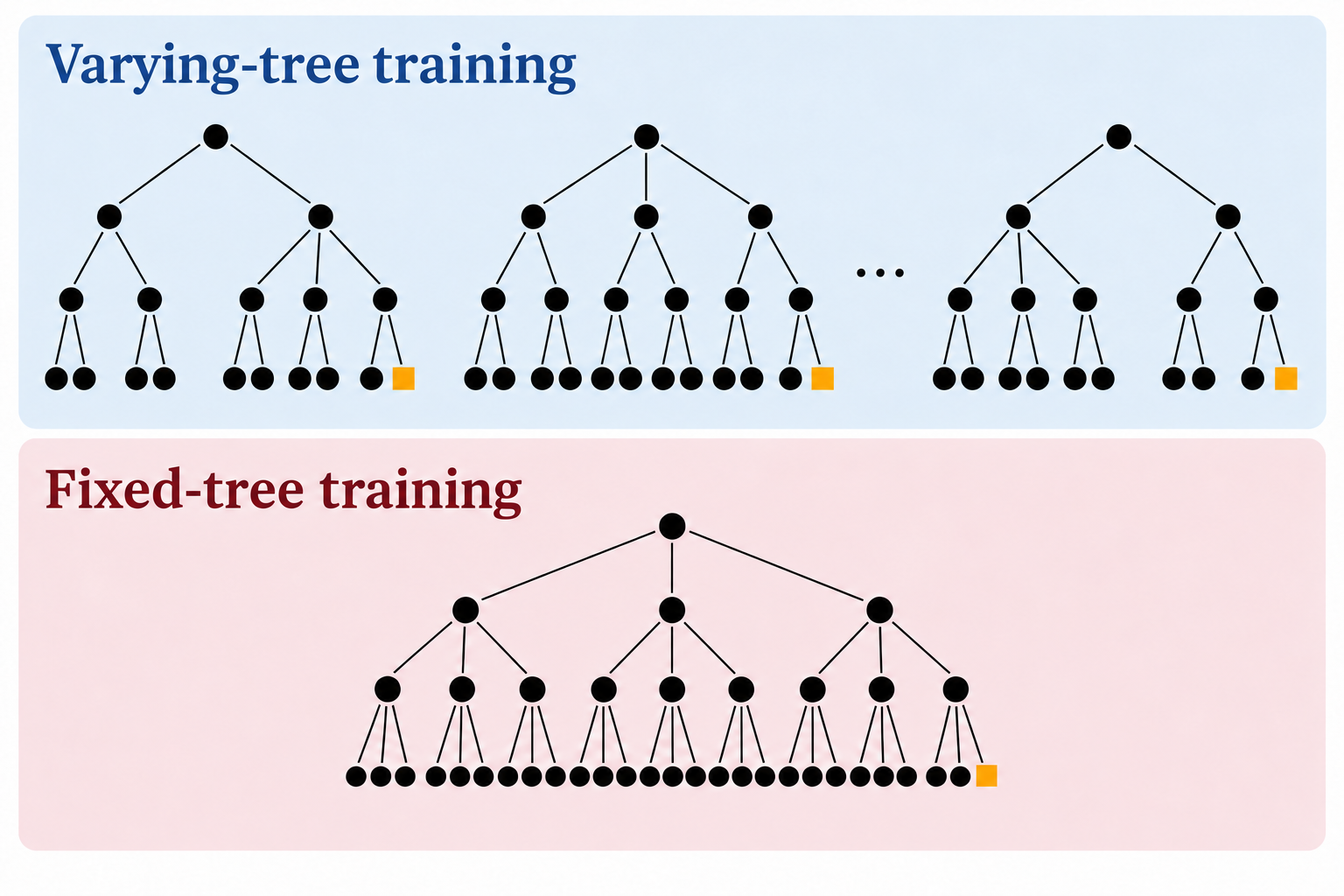}
    \caption{Sketch illustrating our two training protocols. In both cases, we consider a self-supervised sentence completion task, i.e. of predicting $x_d$ given the prefix $x_1,\dots x_{d-1}$. In varying-tree training (top), the sentences can have arbitrary tree topology, and are generated using $p_2{=}p_3{=}1/2$, i.e. equally probable branching ratios. In fixed-tree training, instead (bottom), sentences are derived according to the fixed (ternary) tree topology. For evaluation, we consider in both cases sentence completion on the ternary tree, i.e. the bottom case. }
    \label{fig:varyig_vs_fixed}
\end{figure}

As sketched in Fig.~\ref{fig:varyig_vs_fixed}, we consider the following training/evaluation setup. Our Transformers can be trained on either sentences with arbitrary tree topology from the VT-RHM with $p_2{=}p_3{=}1/2$, i.e. equally probable binary/ternary branchings, or sentences derived only from the strictly ternary fixed-tree topology ($p_3=1$). In both cases, we test for sentence completion on a disjoint set of sentences with ternary fixed-tree topology. In compact notation, therefore, 
\begin{equation}
    \mathbf{X}_{\rm{train}}\sim \{\mathcal{P}_{\rm{VT}}, \ \mathcal{P}_{\rm{ter}} \}, \quad \mathbf{X}_{\rm{test}}\sim \mathcal{P}_{\rm{ter}}
\end{equation}
where $\mathcal{P}_{\rm{VT}}$ and $\mathcal{P}_{\rm{ter}}$ correspond respectively to the sentence distribution induced by setting $(p_2=p_3=1/2)$ and $(p_2=0,p_3=1)$ for a fixed grammar instance. The rationale for evaluating on the ternary test set is, besides benchmarking, the following. Considering the varying-tree dataset as a task mixture, where each task is sentence completion on a different tree topology, the ternary tree is the last and most difficult to generalize to, hence constituting the bottleneck for complete learning. This has two reasons: firstly, simply because of the model definition, the number of ternary rules $m_3$ per symbol is significantly larger than the number of binary rules $m_2$, so that the associated correlations for learning are weaker (as shown for classification in~\citep{parley2026deep}) and the Transformer thus requires more data to acquire these rules. Secondly, in the varying-tree dataset the ternary topology is in fact very rare, with probability $p_3^{13}=2^{-13}\approx 10^{-4}$. We are thus evaluating genuine structural generalization to a practically unseen topology or circuit.

\section{Experiment details}\label{sec:experiment}

We employ GPT-2-style Transformers~\citep{radford2019language}, consisting of $D$ stacked blocks, each comprising causal (i.e. ``decoder-only") multi-head self-attention and a feedforward multilayer perceptron (MLP), with pre-layer normalization and residual connections. The MLPs use ReLU activations. We refer to each such block as a ``layer'' of the Transformer throughout this work, so that a Transformer of depth $D$ consists of $D$ such layers.

All our experiments below are performed for a VT-RHM grammar instance with $L=3$, and grammar parameters $v=8,f=1/4$. To accommodate all sentence lengths, we fix the context window to $T=26=3^L-1$. For the ternary topology, the prefix $x_1,\dots x_{d-1}$ fills up the entire window, while for shorter sentences we simply pad with zeros to the left. We consider a fixed batch size $B=128$, and online learning using Adam~\citep{kingma2015adam}, optimizing over learning rate and weight decay. We further make standard parameter choices for embedding dimension, number of heads and hidden size of the MLP: $d_{\rm emb}=512$ ($256$ for $D=12$), $n_h=16$ and $d_{mlp}=4 \times d_{\rm emb}$. We consider learnable positional embeddings. The last-token prediction is read off from the final position at the last layer, and other positions are not trained on (as they would be for next-token prediction).

\section{Loss stages}\label{sec:stages}

In Fig.~\ref{fig:training_curves}, we show the learning curves for our GPT-2 style Transformers evaluated on the ternary sentence completion task. Strikingly, the last-token cross-entropy test loss shows a clear stagewise behaviour. The key insight in~\citep{cagnetta_how_2024} was to show that such stages can be understood by considering the relative strength of correlation between the last token and preceding segments of the sentence, which can be used to build representations of the hidden latent tree variables. In~\citep{cagnetta_how_2024}, these insights were developed, for fixed-tree training, into a theory of sample complexity (i.e. number of data required) of each learning stage. Here, we will leave open a precise mathematical characterization of sample complexity, but build on these ideas to interpret the varying-tree learning curves: in Sec.~\ref{sec:probing}, we will confirm by probing that each stage indeed can be associated to the emergence of the corresponding hidden variable representation.

\paragraph{Stages 1a and 1b: from bigram/trigram confusion to the fixed-tree trigram.}

In the initial learning stage, the Transformer exploits trigram statistics to predict the last token. For the fixed-tree ($p_2=0$, $p_3=1$), the corresponding loss can be estimated by counting the number of compatible last-token candidates, following~\citep{cagnetta_towards_2024}. For the varying tree, however, the final trigram $(abc)$ can admit two distinct derivations: (i) directly through a ternary rule $(z\to abc)$, or (ii) through a binary rule $(z'\to bc)$, preceded by a subtree whose expansion ends with $a$. When $p_2=p_3=1/2$ and $f_2=f_3=f$, these two configurations contribute comparable probabilities, leading to a higher trigram loss.

To estimate this loss, we account for the number of compatible binary and ternary rules, as well as their possible overlap. Denoting by $N_T$ and $N_B$ the respective numbers of compatible last-token candidates, and by $\mathbb{E}_{\mathcal{G}}[\cdot]$ the expectation over grammar realizations, we have $\mathbb{E}_{\mathcal{G}}[N_T]=\mathbb{E}_{\mathcal{G}}[N_B]=vf$. In the limit $vf\gg1$, these counts concentrate around their mean, yielding the following ensemble-averaged losses, defined as $\mathcal{L}_{trig}=\mathbb{E}_{\mathcal{G}}[H_{\mathcal{G}}(x_d\mid x_{d-2},x_{d-1})]$ (see Appendix~\ref{app:trigram}):
\begin{equation}
\begin{aligned}
\mathcal{L}_{trig}^{FT}
&\simeq \ln(vf),\\
\mathcal{L}_{trig}^{VT}
&\simeq \ln(vf)+(1-f)\ln 2.
\end{aligned}
\label{eq:trigram_losses}
\end{equation}
Thus, the excess loss due to the ambiguity between binary and ternary derivations is approximately $(1-f)\ln 2$. As shown in Fig.~\ref{fig:training_curves}, for varying-tree training the Transformer loss in the initial stage indeed agrees well with our analytical prediction~\footnote{We further test the validity of the large-$vf$ approximation by comparing it with the empirical VT trigram entropy measured by sampling grammars at multiple values of $(v,f)$, see Fig.~\ref{fig:trigram_finite_v}. We show in particular that is already accurate for $v=8,f=1/4$.} for $\mathcal{L}_{trig}^{VT}$: we refer to this as bigram-trigram confusion, and it is due to the fact that the Transformer cannot locally distinguish between the genuine ternary rules, where one should apply trigram statistics, and their fortuitous occurrence as spurious symbol followed by binary rule, where it is optimal to use the bigram.

To progress from the varying-tree to the fixed-tree trigram, the Transformer must learn to distinguish genuine ternary suffixes from fortuitous binary configurations. In particular, if we condition on the sentence ending by two consecutive grammatical triplets, then, neglecting rare $\mathcal{O}(f^3)$ events, the final triplet must arise from a genuine ternary production. Conditioned on this information, the fixed-tree loss $\mathcal{L}_{trig}^{FT}$ becomes attainable. We test this hypothesis through probing in Sec.~\ref{sec:probing}. We show there that there is a transient regime where the penultimate triplet is recognized (i.e. its latent is represented), but not the third-last one: this allows the Transformer to correctly apply the trigram rule on ternary sentences.

\paragraph{Ensuing stages: inside-outside on the depth  $\tell=2,\dots L$ subtree.} 

Beyond Stages 1a and 1b, the learning curves reflect the stagewise learning of the rules. For last-token prediction on the varying-tree, we can (numerically) obtain the Bayes-optimal prediction given knowledge of the grammar rules up to depth $\tilde{L}=2, \dots L$, by running a ``depth-restricted" inside-outside algorithm. This is defined precisely in App.~\ref{app:I-O}, and yields (for $\tilde{L}=2$) the dotted horizontal line shown in Fig.~\ref{fig:training_curves}. 

One may notice that this value is practically indistinguishable from its fixed-tree counterpart, which can be computed analytically following the iterative argument in~\citep{cagnetta_how_2024} (also recalled in App.~\ref{app:last_token_VT}). The intuition behind the argument is that each new layer of grammar rules can be used to constrain the number of compatible parent candidates. In App.~\ref{app:last_token_VT}, where we consider the full range of $f\in[0,1]$, we show that the fixed and varying-tree loss values coincide up to higher-order corrections over a wide range (roughly up to $f\approx 3/8$). This is the regime of $f$ where, although there is local ambiguity, if the grammar rules are known the topology of the sub-tree is uniquely determined (no global ambiguity), as discussed in~\citep{parley2026deep}. The intuition is that, once the topology is uniquely determined, the last-token entropy is simply set by $f$, which we have chosen to be equal for both rule types~\footnote{We do not study the behaviour for $f_2\neq f_3$, which would unnecessarily complicate the analysis.}. We stress that, although the loss values practically coincide numerically, and we denote both of them by the same symbol $\mathcal{L}_{\tilde{L}}$, attaining this value in the varying-tree case still requires a significantly more complex algorithm (inside-outside) to fix the correct tree topology, as we will reveal below by probing (Fig.~\ref{fig:visualization}).

\begin{figure*}[t]
    \centering

    \begin{minipage}[t]{0.48\textwidth}
        \centering
        \includegraphics[
            width=\textwidth
        ]{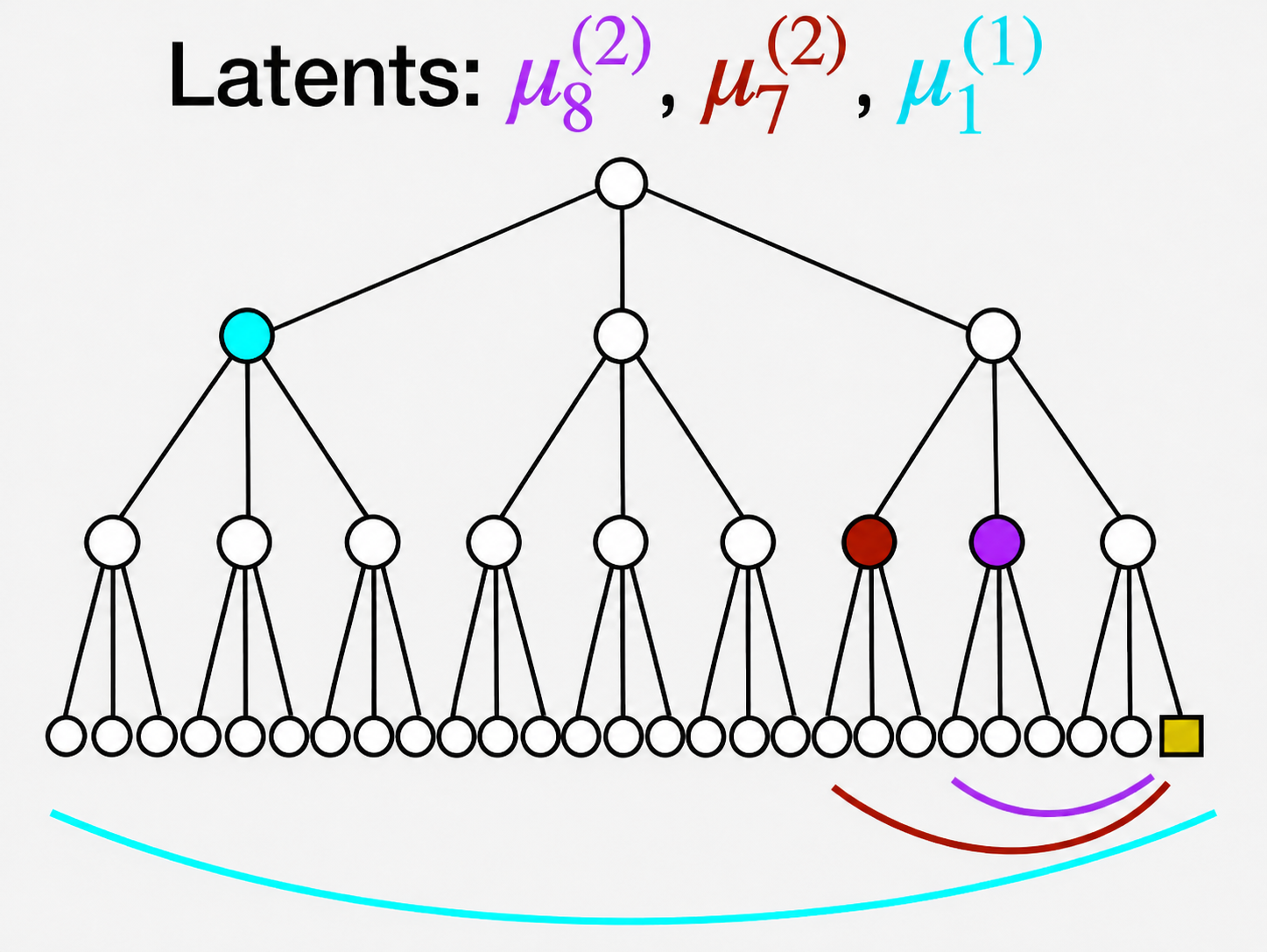}
    \end{minipage}
    \hfill
    \begin{minipage}[t]{0.48\textwidth}
        \centering
        \includegraphics[
            width=\textwidth
        ]{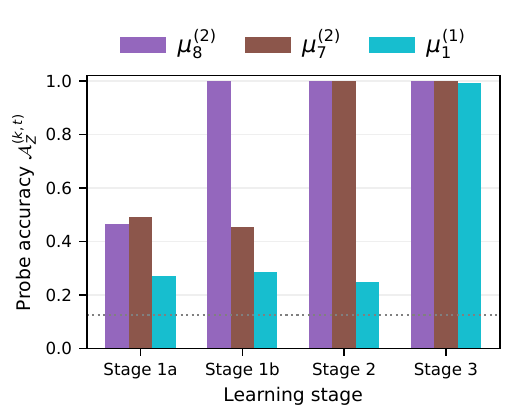}
    \end{minipage}

    \vspace{0.5em}

    \begin{minipage}[t]{0.6\textwidth}
        \centering
        \includegraphics[
            width=\textwidth
        ]{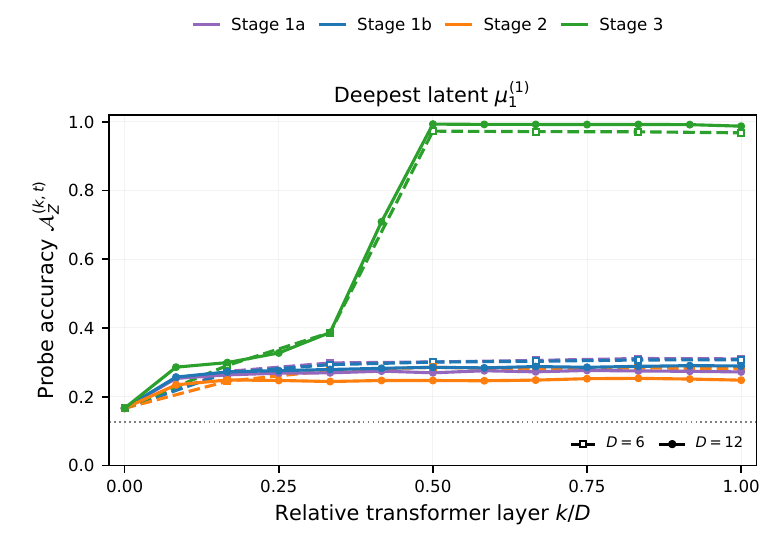}
    \end{minipage}

    \caption{\textbf{Stagewise emegence of hidden variable (nonterminal) representations }
    Top left: sketch of the ternary tree topology, indicating the location of the three hidden latent variables which we focus on for interpretability. Our testable hypothesis is that each successive learning stage of the Transformer will be enabled by linearly representing (in its activations) the sentence ground-truth identity of the corresponding variable, allowing to exploit longer contexts. At each stage, we plot the test accuracy of the trained linear probe for each latent variable (see text), finding a clean correspondence between learning stage and linear representation of the required latent. (c) These probe accuracies saturate already at the middle layer of the networks. We show this in particular for the deepest latent, which is required for the final Stage 3, in Transformers of depth $D=6$ and $D=12$, showing that it is linearly represented beyond the middle layer in both cases.  We refer to App.~\ref{app:supplementary} for the accuracy plots across depth of the other two latents.
    }
    \label{fig:encoder}
\end{figure*}

\begin{figure*}[t]
    \centering
    \includegraphics[
        width=\textwidth
    ]{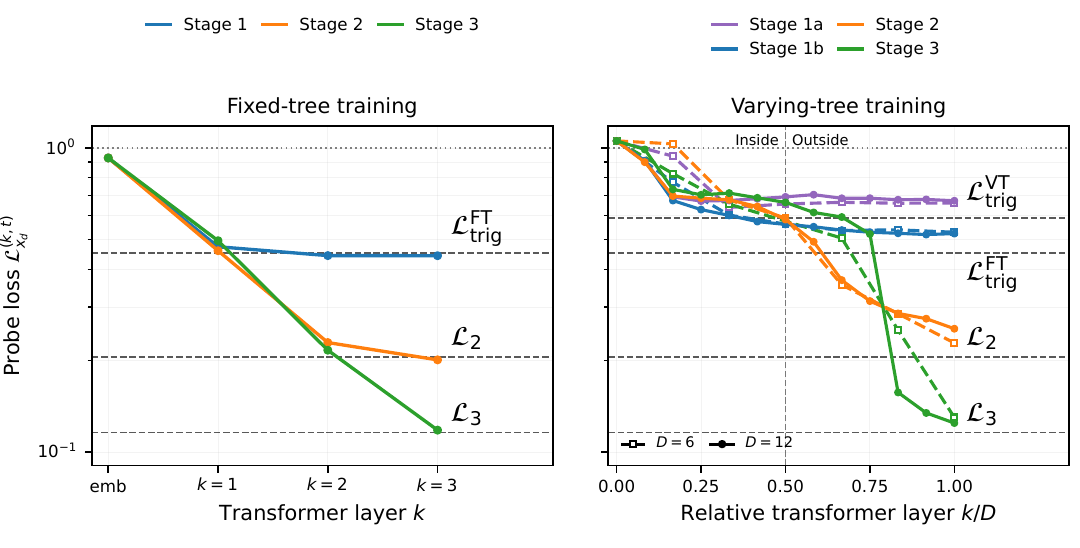}
    \caption{\textbf{Visualization of inside-outside structure and contrast to fixed-tree training.} Left: test cross-entropy loss of the last-token probe across depth for fixed-tree training. We show for simplicity the matched-depth case $D=L=3$, where, strikingly, the Bayes-optimal loss for the $k$-th stage can already be recovered after the $k$-th layer of the Transformer. Right: under varying-tree training, the Transformer organizes itself in a radically different way, as expected from the required inside-outside computation. Our controlled synthetic grammar allows to cleanly visualize this structure: vertical dashed lines indicate the inside-outside boundary at the middle of the network, referring to the fact that the inside information of the most abstract latent has already been encoded (Fig.~\ref{fig:encoder}). The outside pass is implemented as a progressive refinement of the token prediction across the second half of the network, with the later stage (3) requiring more layers to converge to the final Bayes-optimal value. 
    }
    \label{fig:visualization}
\end{figure*}

\section{Interpretability analysis}\label{sec:probing}

To open the door to interpretability, we now perform linear probing experiments on the residual-stream activations of the Transformer. We will first briefly introduce our probing method, and then present its two applications. On the one hand, it will allow us to test our understanding of the learning stages as arising from the representation of deeper and deeper (or more and more abstract) hidden grammatical structure. On the other hand, it will allow us to visualize cleanly the inside-outside organization of computation across depth, which we will do by tracking the improvement of the last-token probe loss.


\textbf{Linear probing methodology.} Consider a Transformer with \(D\) layers and embedding dimension \(d_{\mathrm{emb}}\). Given an input prefix \(\mathbf{x}=(x_1,\ldots,x_{d-1})\), we denote by
\begin{equation}
h_i^{(k,t)}(\mathbf{x})\in\mathbb R^{d_{\mathrm{emb}}}
\end{equation}
the residual-stream representation at position \(i\), after layer \(k\), for the model obtained at training step \(t\). We use \(k=0\) to denote the initial embedding representation, and \(k=1,\ldots,D\) for the outputs of the successive Transformer layers.

For a categorical target variable \(Z\), taking values in \(\{1,\ldots,v\}\), we train a separate linear probe at each residual-stream depth \(k\) and training step \(t\). The probe defines a probability distribution
\begin{equation}
q_Z^{(k,t)}(z\mid \mathbf{x})
=
\operatorname{softmax}
\left(
W_Z^{(k,t)}h_i^{(k,t)}(\mathbf{x})+b_Z^{(k,t)}
\right)_z,
\end{equation}
where $W_Z^{(k,t)}\in\mathbb R^{v \times d_{\mathrm{emb}}}$ and $b_Z^{(k,t)}\in\mathbb R^{v}$ are the probe parameters. The token position \(i\) is chosen according to the target variable, as specified below.

We evaluate the probes on an independent test set, measuring both the classification accuracy
\begin{equation}
\mathcal A_Z^{(k,t)}
=
\mathbb E_{\mathbf{X},Z}\left[
\mathds{1}\left\{
\arg\max_z q_Z^{(k,t)}(z\mid \mathbf{X})=Z
\right\}
\right]
\end{equation}
and, to facilitate comparison with our training curves, the normalized cross-entropy loss
\begin{equation}
\mathcal L_Z^{(k,t)}
=
-\frac{1}{\ln v}
\mathbb E_{\mathbf{X},Z}\left[
\ln q_Z^{(k,t)}(Z\mid \mathbf{X})
\right].
\end{equation}
Here, $\mathbb E_{\mathbf{X},Z}$ denotes the expectation over the joint distribution of test sentences and their associated ground-truth latent labels, for a fixed grammar realization. The normalization is chosen so that uniform predictions correspond to \(\mathcal L_Z^{(k,t)}=1\).

The probes are multinomial logistic classifiers, trained on standardized residual-stream activations. Separate probes are fit independently at different depths and training stages, and regularization is performed in the standard way by controlling for comparable performances on the probe training and test sets.

\textbf{Probing for hidden latent tree variables (nonterminals).} We first investigate whether the Transformer representations encode information about the hidden latent variables of the derivation (parse tree). Let \(\mu_j^{(\ell)}\) denote the latent symbol associated with node \(j\) at grammatical depth \(\ell\), using the root-based indexing convention introduced in Sec.~\ref{sec:VT-RHM}. We focus in particular on three specific latent variables on the ternary tree, whose encoding is essential at each ensuing stage of learning. Namely
\begin{equation}
    \{ \mu_8^{(2)}, \mu_7^{(2)}, \mu_1^{(1)} \}
\end{equation}
which are sketched in Fig.~\ref{fig:encoder}. Building on the observations of~\citep{zhou2026hierarchies,allen-zhu_physics_2023}, we assume that, due to the causal masking of attention, the relevant position to test for encoding of the variable is at the span boundary position, that is $i=24,21$ and $9$ for the three latents respectively.

As representatives of each stage, we consider Transformer model checkpoints at $t=1024, 16384, 262144$ and $2097152$, indicated by the stars in Fig.~\ref{fig:training_curves}. We then create training sets made up of $P_{\rm train}=2^{15}$ residual-stream activations from $P_{\rm train}$ ternary sentences $\mathbf{X}\sim \mathcal{P}_{\rm{ter}}$, and train the probes on predicting the target latent symbol for each sentence.

Our measurements yield the whole profile across depth of the probing accuracy for each latent at each stage: we refer to App.~\ref{app:supplementary} for the full plots. In Fig.~\ref{fig:encoder}, we present the main results. Firstly, we test the correspondence between each learning stage and representation of the required latent, finding a clean agreement (top right). Indeed, for Stage 2 and 3, corresponding to inside-outside on sub-trees of depth 2 and 3, the inside information of level 2 and 1 latent variables respectively is a necessary condition. We find also clearly that, dynamically, the middle level-$2$ latent $\mu_8^{(2)}$ (in purple) appears earlier, explaining the existence of the intermediate Stage 1b where the lower, conditioned (fixed-tree) trigram loss is achieved~\footnote{Note that, in the sketch of Fig.~\ref{fig:encoder}, we have depicted Stage 1b as utilizing the preceding 5 tokens; we stress however that in terms of predictive performance identifying only one of the level-2 latents does not improve upon the fixed-tree trigram loss.}. In terms of the behaviour across depth, the most salient result (bottom) is that the most abstract latent information emerges after the middle layer of the network.


\textbf{Tracking last-token performance.} In the above, we have probed the representation of latent grammatical variables, which are built during the inside pass (recall Fig.~\ref{fig:IO_sketch}). To study the outside pass, we will investigate how this inside information is integrated into the last-position representation which will ultimately carry the token prediction. We therefore fix $i=3^L-1=26$, and train linear probes on the correct target $x_d$ for each prefix. 

Mirroring the two training protocols, for the varying-tree checkpoints we create training sets made up of $P_{\rm train}=2^{15}$ activations to sentences $\mathbf{X}\sim \mathcal P_{\rm VT} $, while for the fixed-tree checkpoints (extracted equivalently at each stage) we consider  $\mathbf{X}\sim \mathcal P_{\rm ter} $. As in the experiments, we test the probes on held-out data from $\mathbf{X} \sim \mathcal P_{\rm ter} $, obtaining the probe cross-entropy loss profile \(\mathcal L_{x_d}^{(k,t)}\). For $k=D$, one expects to recover in each case roughly the Bayes-optimal value of the stage. For $k<D$, instead, what we are doing is equivalent to ``depth pruning": given Transformer activations, we are retraining a new linear+softmax (LSM) head on the last position of a pre-terminal layer, effectively chopping off the later layers.

We show the \(\mathcal L_{x_d}^{(k,t)}\) profiles in Fig.~\ref{fig:visualization}, contrasting fixed-tree and varying-tree training. In the first case, where we can simply match the Transformer depth to the depth of the grammar, we find a very simple structure where the Stage-$k$ Bayes-optimal prediction is already linearly decodable after the $k-$th layer. Under varying-tree training, instead, we recover the expected inside-outside structure, which we can visualize cleanly thanks to our controlled synthetic grammar. We do so again for transformers of depth $D=6$ and $12$. Namely, given the results from Fig.~\ref{fig:encoder} we can draw an inside-outside boundary (vertical dashed line) at the middle layer, where we have found the most abstract latent information to be encoded. We see then that the second half of the network uses the inside information to progressively refine the last-token prediction~\footnote{In App.~\ref{app:supplementary}, we show the same analysis for other tree topologies (besides the ternary one) across the varying-tree set, recovering roughly the same profile.}. As expected from the layerwise structure of the algorithm (see Fig.~\ref{fig:IO_sketch}), we find also that the final Stage 3 requires more layers than Stage 2 to converge to the Bayes-optimal value. At early stages of training instead, where the purely local (trigram) statistics are used, we find indeed that only a few initial transformer layers are involved, with the profile rapidly saturating at the final loss value.

\textbf{Discussion: the question of minimal Transformer depth.} A natural question Fig.~\ref{fig:visualization} may raise is why we have not considered the matched-depth setting $D=L$ also in the right panel. Although we do not provide complete evidence for a rigorous theoretical statement, in practice we have found that even after significantly increasing the width $d_{\rm{emb}}$, training over a long horizon ($t \sim 10^7$ steps), and sweeping over Adam hyperparameters, we have not been able to converge to the final stage Bayes-optimal loss using $D=3$ Transformers (see figure in App.~\ref{app:supplementary}). One can therefore conjecture that $D=4=L+1$ is the minimal depth for successful varying-tree training. On the other hand, we have found that increasing depth further up to $D=6=2L$ leads to better and more stable training, while comparing $D=6$ and $D=12$ the training curves are practically identical suggesting a saturation of the depth advantage beyond $D=6$.

\section{Conclusions and open directions}

How do LLMs build up hierarchical representations of language, and end up mimicking highly non-trivial parsing algorithms from computational linguistics? Inspired by this question, we have analyzed the learning of inside-outside computation by GPT-2 style Transformers on a synthetic, controlled grammar, the VT-RHM~\citep{parley2026deep}. We developed an understanding of the loss stages, and contrasted fixed-tree and varying-tree training protocols, highlighting the stark distinctions in learned computational circuits. Our main interpretability result, captured in Fig.~\ref{fig:visualization}, regards organization of computation across depth: we can cleanly show that, while during early stages of training (bigram/trigram) only initial layers are involved, the Transformer later develops a clear encoder-decoder structure, with the middle layers encoding the most abstract information and the latter half of the network progressively refining the token prediction. This general behavior echoes findings in real-world language models~\citep{tenney2019bert,jawahar2019bert,kervadec2026tracing}, which is remarkable, especially given the fact that the VT-RHM is built from very simple ingredients and allows for analytical understanding of all relevant statistics~\citep{parley2026deep}. We argue that this opens the door to using the VT-RHM for building quantitative descriptions of various phenomena, which we include in the following open directions:

\textbf{Learning algorithm and sample complexity of token prediction.} For the supervised root classification problem~\citep{parley2026deep}, it was shown that, at least in the regime of large vocabulary size and number of rules, the rules of the VT-RHM are provably learned~\footnote{The result of \citep{parley2026deep} concerns convergence of a layerwise method of moments; we note that for the fixed-tree RHM, there are now formal learning results for deep convolutional networks trained layerwise on gradient descent~\citep{ren2026provable}, which have yet to be extended to the VT-RHM.} by a layerwise algorithm based on clustering by correlation, with an associated sample complexity scale (number of data). Extending this to last or next-token prediction, it will be important to understand whether a similar formal learning result can be established, and whether it requires thinking beyond a strictly layerwise algorithm. Furthermore, it will be interesting to characterize precisely which latent-token correlations must be used to build the outside pass (decoder) in a sample-efficient manner.

\textbf{Representation geometry.} Although we have tested the expected correspondence between latent representations and learning stages, we have not studied how the Transformer actually encodes the relevant information. Our controlled grammar may give insight into the origin of syntactic encoding in LLMs: in~\citep{diego2024polar}, it was found that syntactic relations are encoded by the relative direction between embeddings, but it is not understood how such representations are learned.

\textbf{Thresholds to syntactic generalization.} In Fig.~\ref{fig:visualization}, we have only considered the two ``extreme" forms of training, i.e. fixed-tree vs the full varying-tree dataset. A natural question that arises is how the Transformer computation evolves between the two: how rich does the tree mixture need to be in order to impose encoder-decoder structure? This may also shed quantitative insight into phenomena such as Grokking of hierarchical (i.e. tree-like) structure in Transformers~\citep{murty_grokking_2023}.

We hope our work encourages the community to build on this data model and extensions thereof. All relevant code is publicly available on the Github page~\url{https://github.com/jackparley/interpretable_inside_outside}.

\section*{Acknowledgements}
The author thanks Matthieu Wyart and Francesco Cagnetta for important discussions during the early phases of this project. We further acknowledge F.C. as the author of the Github repository which was built upon in order to study the varying-tree model. The author acknowledges funding from the PCSL lab (Prof. M. Wyart) at EPFL, where this research was performed, and the SCITAS infrastructure which was used in order to perform the experiments (mainly H100 GPUs).

\bibliography{Inside_outside}
\bibliographystyle{tmlr}

\newpage
\appendix
\onecolumn


\section{Last-token conditional entropy on the varying-tree}\label{app:last_token_VT}

In this appendix, we show detailed numerical evidence for the claim in the main text, namely that the fixed-tree and varying-tree ensemble-averaged last-token conditional entropy
\begin{equation}
    \mathbb{E}_{\mathcal{G}}[H_{\mathcal{G}}(X_d \mid X_1,\dots,X_{d-1})]
\end{equation}
are practically indistinguishable over a wide range of $f$, up to around $f{\approx} 3/8$.

Before turning to numerics, we recall how the fixed-tree expected losses can be determined following the iterative argument  in~\citep{cagnetta_towards_2024}. Referring to $N_{\tilde L}$ as the number of last-token candidates under a sub-tree of depth $\tell$, we have that~\footnote{Note that we use a different notation to~\citep{cagnetta_towards_2024} for the learning steps. Here, we refer to them by the depth $\tell$ of the subtree on which inside-outside is being performed; in~\citep{cagnetta_towards_2024}, their notation refers to the level of the deepest latents involved.}
\begin{equation}\label{eq:recursion}
\mathbb{E}[N_{\tilde L}]
=
1+(v-1)
\frac{
m_3\,\mathbb{E}[N_{\tilde L-1}]-1
}{
v^3-1
},
\qquad \tilde L\geq2.
\end{equation}
The corresponding loss is then approximated by $\mathcal{L}_{\tilde L}^{FT}\approx
\ln\mathbb{E}[N_{\tilde L}]$, and can be computed iteratively starting from the fixed-tree trigram loss $\mathcal{L}_{trig}^{FT}\equiv\mathcal{L}_{1}^{FT}$ introduced in the main text, which we recall is approximated as $\mathcal{L}_{1}^{FT}\simeq \ln(vf)$ for large $vf$.

\begin{figure}
    \centering
    \includegraphics[width=0.75\linewidth]{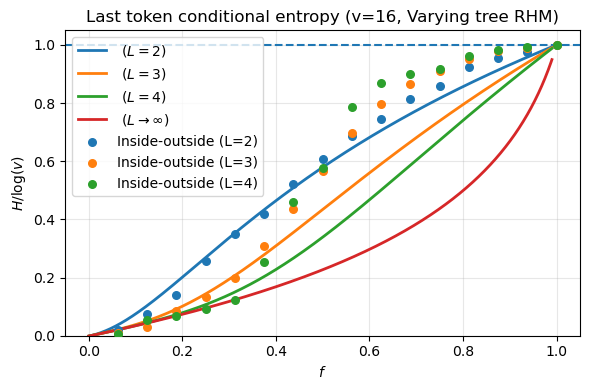}
    \caption{Comparison of the theoretical curves (full lines) obtained from the fixed-tree recursion (\ref{eq:recursion}), with the expected last-token conditional entropies computed numerically using inside-outside on varying-tree grammar datasets. We consider $v=16$, vary $f \in [0,1]$, and grammars of different depth $L$. Up to roughly $f\approx 3/8$, there is good agreement: this corresponds to the regime of no global ambiguity (i.e. single topology per sentence) discussed in~\citep{parley2026deep}.}
    \label{fig:compare_theory}
\end{figure}

In Fig.~\ref{fig:compare_theory}, we show, for fixed $v=16$ and varying $f$, the analytical curves computed from the above recursion (\ref{eq:recursion}). We then sample VT-RHM grammars of the same depth $\tell=L$, to test whether the last-token conditional entropy is well captured by the analytical curves. This is done by sampling many sentences $\mathbf{x}$ from each grammar, running inside-outside (see App.~\ref{app:I-O}), then averaging over both sentences and grammar instances~\footnote{This last averaging is not in practice needed, due to self-averaging even for moderate values of $v$.}. We find that indeed there is good agreement, up to around $f\approx 3/8$. This was the same range of $f$, where, in~\citep{parley2026deep} (see Fig. 2 there), it was shown that the grammars are globally unambiguous, i.e. that the probability of a sentence being derived by two different tree topologies is negligible. We leave open a precise analytical characterization of the varying-tree last-token entropies over the full range of $f\in[0,1]$, which may be possible by considering the additional terms due to multiple tree topologies per sentence.

\section{Tensor formulation of inside-outside}\label{app:I-O}
Here, we provide the detailed equations for the inside-outside algorithm, which for our layered grammars we can formulate in a tensorized fashion: this is crucial for their parallelized implementation (available on Github), which becomes computationally crucial already for grammars of depth $L\geq 3$. Following the notation in~\citep{parley2026deep}, we denote the level-$\ell$ rule tensors by $R^{(\ell)}$. Their
entries are
\begin{equation}
R^{(\ell)}_{z\to(a,b)}
=
\begin{cases}
p_2/m_2, & \text{if } z\to(a,b) \text{ is grammatical},\\
0, & \text{otherwise},
\end{cases}
\end{equation}
for binary productions, and
\begin{equation}
R^{(\ell)}_{z\to(a,b,c)}
=
\begin{cases}
p_3/m_3, & \text{if } z\to(a,b,c) \text{ is grammatical},\\
0, & \text{otherwise},
\end{cases}
\end{equation}
for ternary productions. Here $p_2$ and $p_3$ are the probabilities of
choosing a binary or ternary expansion at a node, respectively, with $p_2+p_3=1$. Conditional on the branching factor, the corresponding grammatical rules
are therefore equiprobable.
In a layered PCFG, given the rule tensors $R^{(\ell)}$, the rule-based parser computes the level-$(\ell-1)$ chart/tensor $M^{(\ell-1)}$ following the recursion

\begin{equation}
\begin{aligned}
&M^{(\ell-1)}_{i,\lambda}[z] =
\sum_{\lambda',a,b}
R^{(\ell)}_{z\ \to\ (a,b)} \times
M^{(\ell)}_{i,\lambda'}[a] \times
M^{(\ell)}_{i+\lambda',\lambda-\lambda'}[b] \\
&\quad + \sum_{\lambda',\lambda'',a,b,c}
R^{(\ell)}_{z\ \to\ (a,b,c)} \times \\
&\qquad M^{(\ell)}_{i,\lambda'}[a] \times
M^{(\ell)}_{i+\lambda',\lambda''}[b] \times
M^{(\ell)}_{i+\lambda'+\lambda'',\lambda-\lambda'-\lambda''}[c].
\end{aligned}
\label{eq:inside_recursion}
\end{equation}

For each prefix $x_1,\dots x_{d-1}$, the chart is initialized at the $L$-th (terminal) level as $M^{(L)}_{i,1}[z]=\delta_{x_i,z}$ for $i=1,\dots d-1$, and $M^{(L)}_{d,1}[z]=1/v, \ \forall z$ at the last position. Above, we have expressed the parser in its sum-product version (inside algorithm), so that the chart stores inside probabilities $
M^{(\ell)}_{i,\lambda}[z]=Pr(x_i,\dots,x_{i+\lambda-1}\mid z)$ for $z \in \mathcal{N}^{(\ell)}$. Alternative choices of operation recover other standard parsing algorithms, such as recognition/CYK~\citep{younger_cyk_1967}.  

The inside algorithm computes, for each symbol and span, the probability of generating the substring inside that span. For last-token prediction, we instead need the complementary quantity: the probability of generating the observed context outside a given span. This is computed by the corresponding outside pass.

The outside chart $O$ is defined so that
\begin{equation}  
    O^{(\ell)}_{i,\lambda}[z]
\end{equation}
stores the total weight of all derivations of the observed sentence outside the span
$(i,\lambda)$, conditional on this span being generated by the symbol
$z\in \mathcal{N}^{(\ell)}$.

The outside pass is, in general, initialized at the root level (see note below on marginalizing over the root symbol) by
\begin{equation}
    O^{(0)}_{1,d}[y] = p(y),
    \qquad y\in \mathcal{N}^{(0)},
\end{equation}
It is then computed recursively from level
$\ell-1$ to level $\ell$ by distributing the outside weight of each parent span to its children.

For a binary rule $z\to(a,b)$, each decomposition of a span $(i,\lambda)$ into
$(i,\lambda')$ and $(i+\lambda',\lambda-\lambda')$ contributes
\begin{equation}
\begin{aligned}
O^{(\ell)}_{i,\lambda'}[a]
&\mathrel{+}=
O^{(\ell-1)}_{i,\lambda}[z]\,
R^{(\ell)}_{z\to(a,b)}\,
M^{(\ell)}_{i+\lambda',\lambda-\lambda'}[b],
\\
O^{(\ell)}_{i+\lambda',\lambda-\lambda'}[b]
&\mathrel{+}=
O^{(\ell-1)}_{i,\lambda}[z]\,
R^{(\ell)}_{z\to(a,b)}\,
M^{(\ell)}_{i,\lambda'}[a].
\end{aligned}
\end{equation}
Similarly, for a ternary rule $z\to(a,b,c)$, each decomposition into spans
$(i,\lambda')$, $(i+\lambda',\lambda'')$, and
$(i+\lambda'+\lambda'',\lambda-\lambda'-\lambda'')$ contributes
\begin{equation}
\begin{aligned}
O^{(\ell)}_{i,\lambda'}[a]
&\mathrel{+}=
O^{(\ell-1)}_{i,\lambda}[z]\,
R^{(\ell)}_{z\to(a,b,c)}\,
M^{(\ell)}_{i+\lambda',\lambda''}[b]\,
M^{(\ell)}_{i+\lambda'+\lambda'',\lambda-\lambda'-\lambda''}[c],
\\
O^{(\ell)}_{i+\lambda',\lambda''}[b]
&\mathrel{+}=
O^{(\ell-1)}_{i,\lambda}[z]\,
R^{(\ell)}_{z\to(a,b,c)}\,
M^{(\ell)}_{i,\lambda'}[a]\,
M^{(\ell)}_{i+\lambda'+\lambda'',\lambda-\lambda'-\lambda''}[c],
\\
O^{(\ell)}_{i+\lambda'+\lambda'',\lambda-\lambda'-\lambda''}[c]
&\mathrel{+}=
O^{(\ell-1)}_{i,\lambda}[z]\,
R^{(\ell)}_{z\to(a,b,c)}\,
M^{(\ell)}_{i,\lambda'}[a]\,
M^{(\ell)}_{i+\lambda',\lambda''}[b].
\end{aligned}
\end{equation}

The outside chart can be used to compute posterior marginals over any span. Indeed, for any span $(i,\lambda)$ and symbol $z\in\mathcal{N}^{(\ell)}$,
\begin{equation}
Pr\!\left(z \Rightarrow x_i,\dots,x_{i+\lambda-1}\mid \mathbf{x}\right)
=
\frac{
O^{(\ell)}_{i,\lambda}[z]\,
M^{(\ell)}_{i,\lambda}[z]
}{
Pr(\mathbf{x})
},
\end{equation}
where
\begin{equation}
Pr(\mathbf{x})
=
\sum_{y\in\mathcal{N}^{(0)}}
p(y)\,M^{(0)}_{1,d}[y].
\end{equation}

For last-token prediction, the context
$(x_1,\dots,x_{d-1})$ is observed and the final token $x_d$ is masked. After running the outside pass, the outside values
\begin{equation}
O^{(L)}_{d,1}[z],
\qquad z\in\mathcal{N}^{(L)},
\end{equation}
give the total probability of generating the observed context under the constraint that the final terminal is $z$. Therefore the Bayes-optimal last-token predictor is
\begin{equation}
\hat{x}_d
=
\arg\max_{z\in\mathcal{N}^{(L)}}
O^{(L)}_{d,1}[z],
\end{equation}
and the full predictive distribution is
\begin{equation}
Pr(X_d=z\mid x_1,\dots,x_{d-1})
=
\frac{
O^{(L)}_{d,1}[z]
}{
\sum_{z'\in\mathcal{N}^{(L)}} O^{(L)}_{d,1}[z']
}.
\end{equation}

\subsection{Marginalizing over the root symbol}
For the last-token prediction task, the root class variable can be marginalized out explicitly. Defining the effective root production probabilities
\begin{equation}
\overline R^{(1)}_{a,b}
=\sum_y p(y)R^{(1)}_{y\to(a,b)},
\qquad
\overline R^{(1)}_{a,b,c}
=\sum_y p(y)R^{(1)}_{y\to(a,b,c)},
\end{equation}
the original grammar is equivalent, at the level of the marginal distribution over terminal sequences, to a grammar with a single root symbol and production probabilities \(\overline R^{(1)}\).

Defining the effective root production probabilities
\begin{equation}
\overline R^{(1)}_{a,b}
=\sum_y p(y)R^{(1)}_{y\to(a,b)},
\qquad
\overline R^{(1)}_{a,b,c}
=\sum_y p(y)R^{(1)}_{y\to(a,b,c)},
\end{equation}
the inside pass need only be computed up to level $\ell=1$.
The outside pass can then be initialized directly at this level,
without explicitly constructing the level-0 inside chart or
retaining the root class labels. Furthermore, since we are
only interested in the last token, we need only consider spans
ending at position $d$, for which
\begin{equation}
\boxed{
\begin{aligned}
O^{(1)}_{i,d-i+1}[z]
={}&
\sum_a
\overline R^{(1)}_{a,z}
M^{(1)}_{1,i-1}[a]
\\
&+
\sum_{j=1}^{i-2}\sum_{a,b}
\overline R^{(1)}_{a,b,z}
M^{(1)}_{1,j}[a]
M^{(1)}_{j+1,i-j-1}[b].
\end{aligned}
}
\end{equation}
When predicting only the final token, the outside recursion can be restricted to spans ending at the final position, since the final token necessarily belongs to the rightmost child at every branching. This yields exactly the same Bayes-optimal predictive distribution as the full inside–outside algorithm.

\subsection{Depth-restricted inside-outside}

We also consider a depth-restricted version of the inside--outside
algorithm. Let $\widetilde{L}<L$ denote a cutoff depth. Up to depth
$\widetilde{L}$, inference uses the rule tensors of the particular realized
grammar. At deeper levels, the realization-specific rule tensors are
replaced by their uniform average $\overline{R}$ over rule sampling, that is $p_2/v^2$ across all binary tuples and $p_3/v^3$ across all ternary tuples. Note that, although such rules are almost structureless, this still implies an implicit prior on the fact that there are binary and ternary rules, and they have equal overall  probability weight $p_2=p_3$. We expect the effect of this to be negligible, and we have checked (see e.g. dotted line in Fig.~\ref{fig:training_curves}) that numerically the approach reproduces well the expected last-token entropies.

More precisely, if $R^{(\ell)}$ denotes the rule tensor connecting levels
$\ell-1$ and $\ell$, we define
\begin{equation}
\widetilde{R}^{(\ell)}_{\widetilde{L}}
=
\begin{cases}
R^{(\ell)},
& \ell> L-\widetilde{L},
\\[3pt]
\overline{R}
& \ell\leq L-  \widetilde{L}.
\end{cases}
\end{equation}
The depth-$\widetilde{L}$ inside--outside algorithm is then obtained by
running the usual inside and outside recursions after replacing
$R^{(\ell)}$ by $\widetilde{R}^{(\ell)}_{\widetilde{L}}$ at every level. The case $\widetilde{L}=L$ recovers
the full inside--outside algorithm.

\section{Trigram loss in the varying-tree grammar}
\label{app:trigram}

We derive the large-$vf$ prediction for the conditional entropy
$H(x_d\mid x_{d-2},x_{d-1})$, corresponding to the initial
trigram-learning stage. We consider a grammar with $v$ symbols
per layer, $m_2=fv$ binary rules and $m_3=fv^2$ ternary rules
per parent, and $p_2=p_3=1/2$. The right-hand sides of each
rule type are sampled uniformly without replacement across
all parents. Thus, there are $vm_2=fv^2$ distinct binary
right-hand sides and $vm_3=fv^3$ distinct ternary ones.

\paragraph{Compatible derivations.}

\begin{figure}[t]
    \centering
    \includegraphics[width=0.75\linewidth]{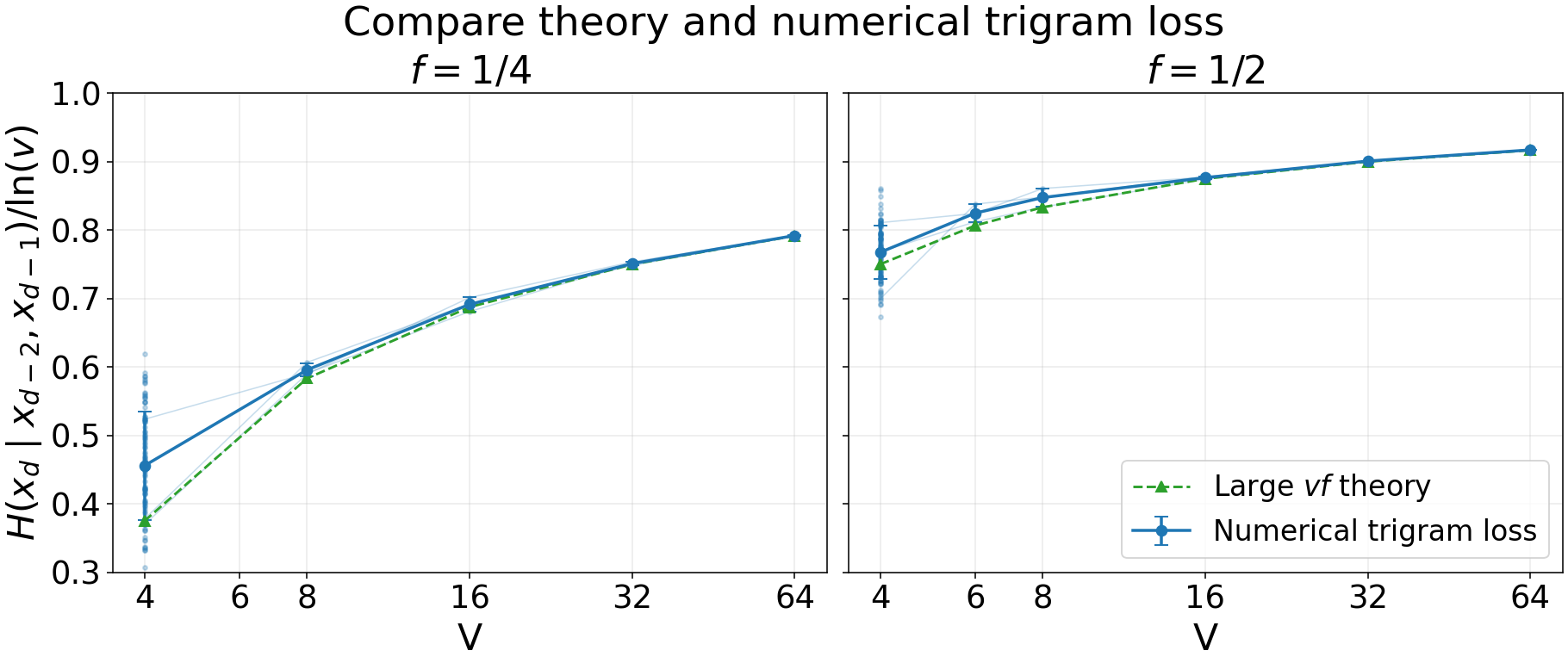}
    \caption{Numerical validation of the large-$vf$ theoretical
    prediction for the varying-tree trigram loss. Sampling
    grammars for various combinations of $(v,f)$, we numerically
    estimate $H(x_d\mid x_{d-2},x_{d-1})$ by averaging over
    sentences. Different grammar realizations (faint curves)
    concentrate around the mean due to self-averaging.
    We compare the numerical results with the theoretical
    prediction $\mathcal L_{trig}^{VT}\simeq
    \ln(vf)+(1-f)\ln2$, normalized by $\ln v$.
    This large $vf$ approximation
    is already very accurate for $v=8, f=1/4$.}
    \label{fig:trigram_finite_v}
\end{figure}

Consider a final trigram $(abc)$. We distinguish two
possible local configurations: a ternary production
$z\to abc$, and a binary production $z'\to bc$ preceded
by a subtree ending with $a$. We approximate the two
configurations as contributing equally weighted compatible
derivations, after marginalizing over the preceding tree.

Define the indicators
\begin{equation}
T_{abc}=\mathds{1}\{abc\text{ is a ternary RHS}\},
\qquad
B_{bc}=\mathds{1}\{bc\text{ is a binary RHS}\}.
\end{equation}
Under the equal-weight approximation, the unnormalized
probability of the trigram is
\begin{equation}
W_{abc}=T_{abc}+B_{bc}.
\end{equation}
In particular, a token compatible with both configurations
receives twice the weight of one compatible with only one.

For a fixed context $(ab)$, define
\begin{equation}
N_T(ab)=\sum_c T_{abc},
\qquad
N_B(b)=\sum_c B_{bc},
\end{equation}
and
\begin{equation}
Z_{ab}=\sum_c W_{abc}=N_T(ab)+N_B(b).
\end{equation}
The conditional distribution of the final token is then
approximated by
\begin{equation}
P(c\mid a,b,\mathcal{G})\simeq\frac{W_{abc}}{Z_{ab}},
\end{equation}
for contexts with $Z_{ab}>0$.

\paragraph{Conditional entropy and overlap correction.}

For a given context $(ab)$, the conditional entropy reads
\begin{align}
H_{ab}
&=-\sum_c\frac{W_{abc}}{Z_{ab}}
\ln\frac{W_{abc}}{Z_{ab}}\nonumber\\
&=\ln Z_{ab}
-\frac{1}{Z_{ab}}\sum_c W_{abc}\ln W_{abc}.
\label{eq:trig_entropy_context}
\end{align}
Since $T_{abc},B_{bc}\in\{0,1\}$, the weights $W_{abc}$
take values in $\{0,1,2\}$. Consequently,
\begin{equation}
W_{abc}\ln W_{abc}
=(T_{abc}B_{bc})\ 2\ln2 .
\end{equation}
given that $W_{abc}\ln W_{abc}$ is only non-zero when both indicators equal 1 (we use the standard convention $0\ln0=0$).
Defining the number of overlapping candidates as
\begin{equation}
D_{ab}=\sum_c T_{abc}B_{bc},
\end{equation}
Eq. (\ref{eq:trig_entropy_context}) becomes
\begin{equation}
H_{ab}=\ln Z_{ab}
-\frac{2D_{ab}}{Z_{ab}}\ln2.
\label{eq:trig_entropy_overlap}
\end{equation}

To obtain the total trigram loss, we must average over
contexts weighted by their probability of occurrence.
Since each allowed ternary RHS contributes once to
$\sum_{abc}W_{abc}$, while each allowed binary RHS
contributes for all $v$ possible values of $a$, we have
\begin{equation}
\sum_{a,b}Z_{ab}
=vm_3+v(vm_2)=2fv^3.
\end{equation}
Thus,
\begin{equation}
P(a,b\mid \mathcal{G})\simeq\frac{Z_{ab}}{2fv^3},
\end{equation}
and the trigram entropy for a fixed grammar is
\begin{align}
\mathcal L_{trig}^{VT}(\mathcal{G})
&=\sum_{a,b}P(a,b\mid \mathcal{G})H_{ab}\nonumber\\
&\simeq\frac{1}{2fv^3}
\sum_{a,b}
\left[Z_{ab}\ln Z_{ab}-2D_{ab}\ln2\right].
\label{eq:trig_grammar_entropy}
\end{align}

We now average over grammar realizations. By symmetry,
the distribution of $Z_{ab}$ is the same for every
context, so that
\begin{equation}
\mathbb E_{\mathcal{G}}\left[\sum_{a,b}Z_{ab}\ln Z_{ab}\right]
=v^2\mathbb E[Z\ln Z],
\end{equation}
where $Z=N_T+N_B$ denotes the count for an arbitrary
fixed context.

Furthermore, binary and ternary rule sets are sampled
independently, with each possible RHS having marginal
inclusion probability $f$. Hence,
\begin{equation}
\mathbb E_{\mathcal{G}}[D_{ab}]
=\sum_c\mathbb E[T_{abc}B_{bc}]
=vf^2.
\end{equation}
Substituting into Eq. (\ref{eq:trig_grammar_entropy}),
we obtain
\begin{equation}
\boxed{
\mathcal L_{trig}^{VT}
\simeq
\frac{\mathbb E[Z\ln Z]}{2vf}
-f\ln2.
}
\label{eq:trig_vt_finite}
\end{equation}
The second term accounts for the overlap between binary
and ternary derivations: two distinct derivations may
produce the same last token, thereby reducing the entropy
relative to treating them as distinct outcomes.

\paragraph{Large-$vf$ approximation.}

The numbers of compatible ternary and binary continuations
are respectively $N_T(ab)=\sum_c T_{abc}$ and
$N_B(b)=\sum_c B_{bc}$. Since each possible right-hand
side is selected with probability $f$, their expectations
over grammar realizations are
\begin{equation}
\mathbb E[N_T]=\mathbb E[N_B]=vf.
\end{equation}
For $vf\gg1$, these counts concentrate around their
mean, so that $Z=N_T+N_B\simeq2vf$. Substituting into
Eq. (\ref{eq:trig_vt_finite}) gives
\begin{equation}
\boxed{
\mathcal L_{trig}^{VT}
\simeq\ln(2vf)-f\ln2
=\ln(vf)+(1-f)\ln2.
}
\label{eq:large_vf_approx}
\end{equation}

For the fixed tree, only ternary derivations contribute,
so that $W_{abc}=T_{abc}$ and $Z_{ab}=N_T(ab)$.
All compatible last tokens therefore have equal
probability. Repeating the context-weighted average,
we obtain
\begin{equation}
\mathcal L_{trig}^{FT}
=\frac{\mathbb E[N_T\ln N_T]}{vf}
\simeq\ln(vf),
\end{equation}
where the final approximation again follows from the
concentration of $N_T$ around its mean. Consequently,
the excess varying-tree trigram loss approaches
\begin{equation}
\boxed{
\mathcal L_{trig}^{VT}-\mathcal L_{trig}^{FT}
\simeq(1-f)\ln2.
}
\label{eq:trig_large_v}
\end{equation}

In Fig.~\ref{fig:trigram_finite_v}, we compare the
large-$vf$ prediction with the trigram conditional
entropies measured numerically by building varying-tree grammar instances and averaging over sentence endings.
The agreement is already very good for $v\geq 8$, confirming
that the large $vf$ prediction already captures well the initial varying-tree trigram loss in the training curves.

\section{Supplementary figures}
\label{app:supplementary}
\begin{figure}
    \centering
    \includegraphics[width=0.8\linewidth]{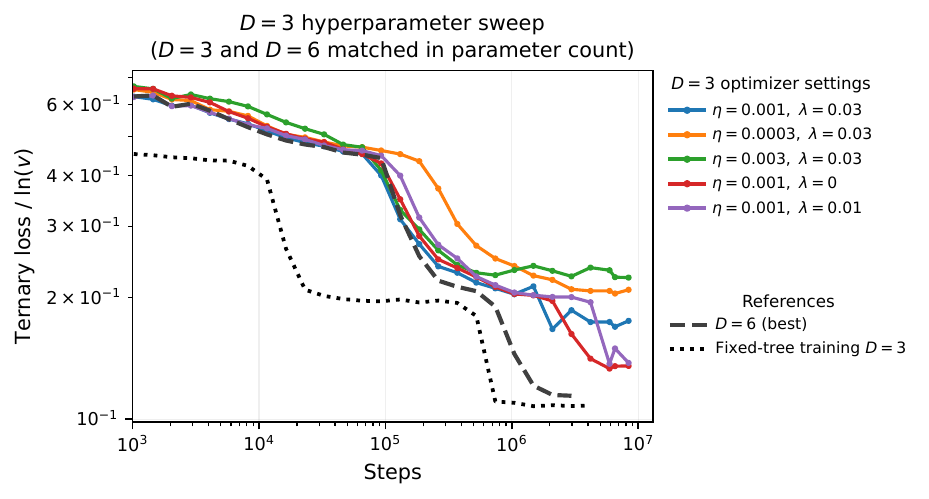}
    \caption{Evaluation loss on the ternary sentence completion task, for depth $D=3$ Transformers. To have roughly the same number of parameters as ($D=6$, $d_{\rm emb}=512$), we increase the embedding dimension to $d_{\rm emb}=768$. Sweeping over Adam hyperparameter settings (learning rate $\eta$, weight decay $\lambda$), we have not managed to achieve full convergence to the final Bayes-optimal loss. For reference, we include the $D=6$ varying-tree training curve as well as the $D=3$ fixed-tree training curve, which converges excellently. }
    \label{fig:D_3_curves}
\end{figure}

We include here some supplementary empirical data which were alluded to in the main text. Firstly, in Fig.~\ref{fig:D_3_curves} we show the training curves for $D=3$ GPT-2 style transformers in the varying-tree setting. As mentioned in the main text, even increasing significantly $d_{\rm emb}$, so that the parameter count matches the $D=6$ model (this count grows quadratically with the embedding dimension), and training over a long window (up to $t \approx 10^7$), we have not converged to the final Bayes-optimal loss.

\begin{figure}
    \centering
    \includegraphics[width=0.68\linewidth]{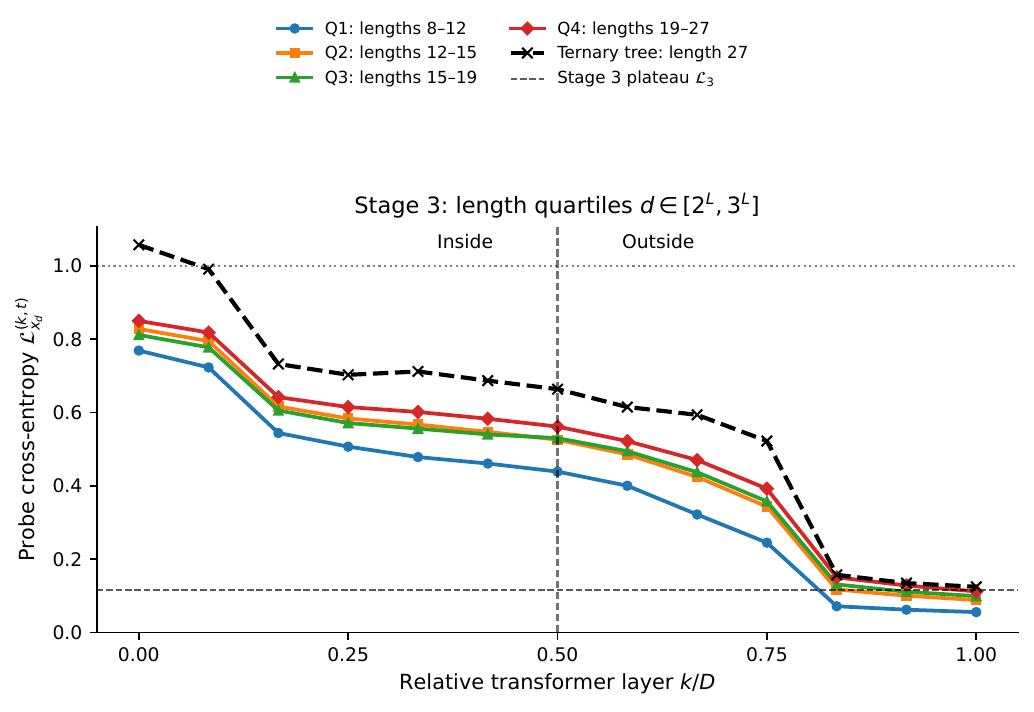}
    \caption{Same analysis as Fig.~\ref{fig:visualization}, but testing the probes on data from various tree topologies across the varying-tree set. For simplicity, we group these according to quartiles of the distribution of sequence lengths. We recover roughly the same profile in all cases. For Q1, corresponding to the shortest sequences, we find some fluctuation in the final plateau value with respect to the ensemble average prediction (this is due to the overwhelming presence of binary rules, for which $m_2=2$ is small). Shown is data for the $D=12$ Transformer at the final Stage 3.}
    \label{fig:quartiles}
\end{figure}

Secondly, Fig.~\ref{fig:quartiles} shows the same analysis as Fig.~\ref{fig:visualization} for Stage 3 (fully trained Transformer), but where the probe test set is composed of other tree topologies, i.e. not just the ternary tree. In particular, we split the probe test data according to the distribution of sequence lengths, defining the quartiles of this distribution. We find roughly the same profile across depth, with the latter half of the network progressively refining the final prediction.

\begin{figure}
    \centering
    \includegraphics[width=0.99\linewidth]{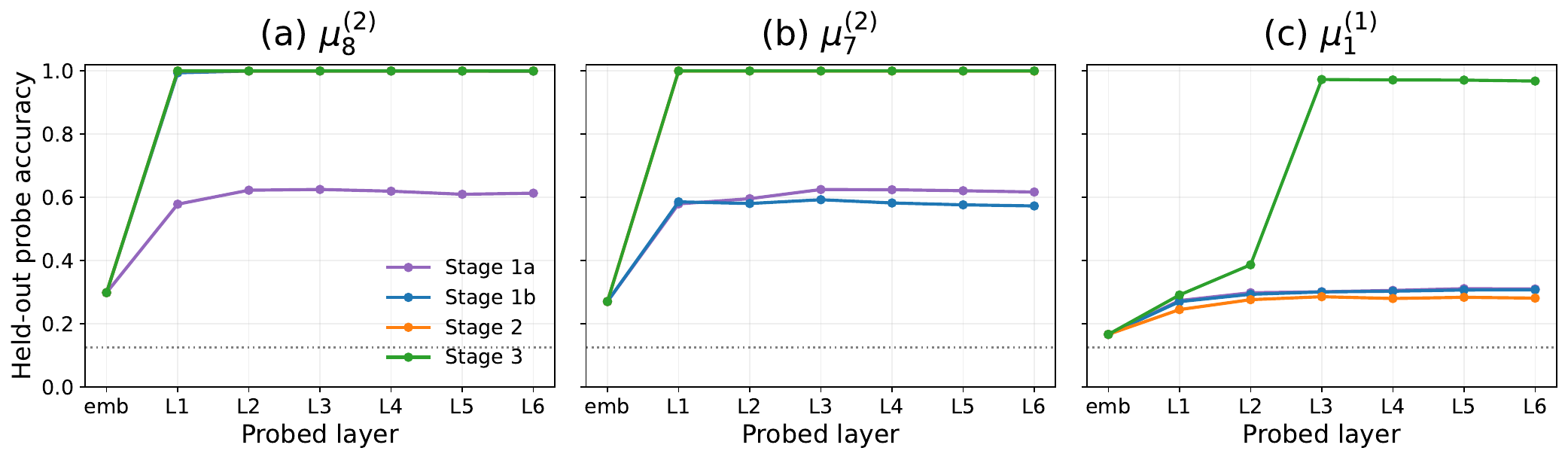}
    \caption{Held-out probe accuracy $\mathcal A_Z^{(k,t)}$ across layers for a $D=6$ Transformer, for the three latents studied in the main text, at the 4 stage checkpoints.}
    \label{fig:three_panel_6}
\end{figure}

\begin{figure}
    \centering
    \includegraphics[width=0.99\linewidth]{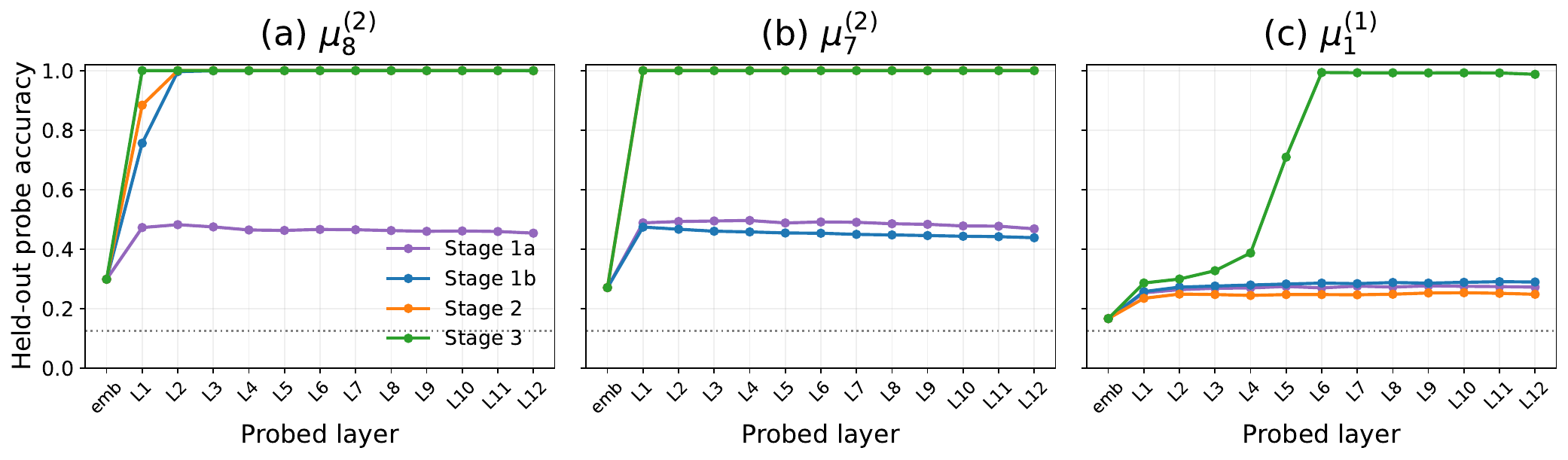}
    \caption{Same as Fig.~\ref{fig:three_panel_6}, but for a $D=12$ Transformer.}
    \label{fig:three_panel_12}
\end{figure}

Finally, in Figs.~\ref{fig:three_panel_6} and \ref{fig:three_panel_12}, we show the full held-out accuracy $\mathcal A_Z^{(k,t)}$ profile across depth for all latents at all stages. These data are used for Fig.~\ref{fig:encoder} (top right); specifically, we extract the accuracy value for each latent at each stage from the middle layer, where it has already saturated (any layer beyond the middle one would give the same result). We do this for $D=12$, although the $D=6$ data show the same stagewise behaviour.

\end{document}